\documentclass[12pt,a4paper]{article}

\usepackage[cmex10,intlimits]{amsmath}
\usepackage{wasysym}
\usepackage{amssymb}
\usepackage{amsthm}
\usepackage{amsfonts}
\usepackage{amsmath}
\usepackage{color}
\usepackage{a4wide}
\usepackage{rotating}
\usepackage{cite}
\usepackage{tikz}
\usepackage{bm}
\usepackage{tipa}
\usepackage[utf8]{inputenc}
\usepackage[T1]{fontenc}

\newcommand{\sign}{\mathrm{sign}}

\makeatletter
\def\blfootnote{\xdef\@thefnmark{}\@footnotetext}
\makeatother

\title{Supervised Classification: Quite a Brief Overview}

\author{Marco~Loog \medskip \\
\small
\begin{tabular}{r|l}
Pattern Recognition Laboratory & The Image Section \\
Delft University of Technology & University of Copenhagen \\
The Netherlands & Denmark \\
e-mail: m.loog@tudelft.nl & http: prlab.tudelft.nl
\end{tabular}}

\begin{document}

\date{August 11, 2017}

\maketitle

\begin{abstract}
The original problem of supervised classification considers the task of automatically assigning objects to their respective classes on the basis of numerical measurements derived from these objects.  Classifiers are the tools that implement the actual functional mapping from these measurements---also called features or inputs---to the so-called class label---or output.  The fields of pattern recognition and machine learning study ways of constructing such classifiers.  The main idea behind supervised methods is that of learning from examples: given a number of example input-output relations, to what extent can the general mapping be learned that takes any new and unseen feature vector to its correct class?  This chapter provides a basic introduction to the underlying ideas of how to come to a supervised classification problem. In addition, it provides an overview of some specific classification techniques, delves into the issues of object representation and classifier evaluation, and (very) briefly covers some variations on the basic supervised classification task that may also be of interest to the practitioner.  \medskip 
\end{abstract}

\newpage

\tableofcontents

\section{An Introduction}

Consider the playful, yet in a way realistic problem of comparing apples and oranges.  More precisely, let us consider the goal of telling them apart in an automated way, e.g. by means of a machine or, more specifically, a computer.  How does one go about building such a machine?  One approach could be to try and construct an accurate physiological model of both types of fruit. We gather what is known about their appearances, their cell structure, their chemical compounds, etc., together with any type of physiological laws that relate these quantities and their inner processes.  Being presented by a new piece of fruit, we can then measure the quantities we think matter most and check these with our two models.  A new piece of fruit for which we want to predict whether it is an apple or an orange is then best assigned to the class for which the model fits best.  Clearly, the performance of this approach critically depends on issues such as how well we can build such models, how good we are at deciding what are the quantities that matter and how accurate we can measure these, how we actually decide whether we have good model fit, and so on.  Such a model may, for instance, not be adequate at all when dealing with cases that are pathological from a physiological point of view, e.g. pieces of fruit that suffer from deformations or rot.  Nevertheless, having enough understanding of the problem at hand, we may be able to tackle it in the way sketched.

\subsection{Learning, not Modelling}

Now, let us consider the, in a way, even more challenging problem of comparing foos and bars.  Again, let us consider the goal of telling them apart in an automated way on the basis of particular measurements taken from the individual foos and bars.  How are we going to go about our problem now?  We might not even know exactly what we are dealing with here. Foos?  Bars?  So, where for the fruits we could try and build, for instance, a physiological model, we now do not even know whether physiology at all applies.  Or is it physics that we need in this case to describe our objects?  Chemistry maybe?  Economics?  Linguistics?

In the absence of any precise knowledge of our problem at hand\footnote{Unfortunately, people may often not be able to recognize such absence. Work by Tversky, Kahneman, and others tells us, for instance, that people can suffer from systematic deviations from rationality or good judgment, i.e., cognitive biases \cite{kahnemann1982}.}, another approach to construct the asked-for machine that tells two or more object classes apart is by means of learning from examples.  With this, we move away from any precisely interpretable model based on more or less factual knowledge about the object classes we are dealing with.  As a substitute, we consider a different type of model---in a sense more general purpose---that can learn from examples.  This is one of the premier learning settings that is studied in the fields of pattern recognition and machine learning\footnote{We do not care to elaborate on the possible distinctions between machine learning and pattern recognition.  For this chapter, it is perfectly fine to consider them one and the same and we, indeed, are going to treat them as such.}.  The particular task is referred to as the supervised classification task: given a number of example input-output relations, can a general mapping be learned that takes any new and unseen feature vector to its correct class?  That is: can we generalize from finite examples?

\subsection{An Outline}

The next section starts out with a basic introduction into classification technology and some of the underlying ideas.  Classifiers are the aforementioned mappings, or functions, that take the measurements made on our objects that need assigning and aim to map these values to the correct corresponding output, or label as it is often referred to.  Section \ref{sect:featrep} discusses a matter that typically warrants consideration prior even to the actual learning, or training, of the classifier: what measurements is the classifier that we are about to construct actually going to rely on when predicting the labels for new and unseen examples?  In the same section, we more broadly cover the issue of object representation, which generally considers in what way to present our objects to the classifier.  At this point, we can build various classifiers and have different possibilities to represent our objects, but really how good is the actual classifier built at our prediction task?  Section \ref{sect:eval} delves into the question of classifier evaluation, providing some tools to get insight into the behavior and performance of the machines that we have constructed.  Two of the chapter's main conclusions are covered in this section: 1) there is no single best classifier and 2) there is an inherent tradeoff between the number of examples from which to generalize and classifier complexity.  Section \ref{sect:reg} then quickly introduces elementary regularization, i.e., another way of controlling classifier complexity.  In conclusion, Section \ref{sect:disc} mentions some variations to basic supervised classification.

\section{Classifiers}\label{sect:clsf}

Before we get to our selection of classifiers, we introduce some notation, make mathematically more precise what the scope is within which we operate, and describe what is the ultimate aim of a classifier in equally unambiguous terms.

\subsection{Preliminaries}\label{sect:pre}

In the general classification setting considered, there are a total of $N$ labeled objects $o_i$, with $i \in \{1,\ldots,N\}$, to learn from.  Every object $o_i$ is represented by a $d$-dimensional feature vector $x_i \in \mathbb{R}^d$ and has corresponding labels $y_i \in \{-1,+1\}$.  So we assume we actually already have a numerical representation that makes up the $d$ measurements in every vector $x_i$.  In addition, with the choice of $y_i \in \{-1,+1\}$, we limit ourselves here to two-class classification problems for simplicity of exposition\footnote{Every two-class classifier can, however, in a more or less principled way, be extended to the general multi-class case. One way is through combining multiple two-class classifiers (see, for instance,\cite{galar2011o}).  Other classifiers, in particular probabilistic ones, adapt more naturally to the multi-class case.}.  Although typically not stated explicitly, we should be aware that the pairs $(x_i,y_i)$ are assumed to be random yet i.i.d.\ draws from some underlying true distribution $p_{XY}$.  For notational convenience, we also define $N_+$ and $N_-$ ($N_+ +N_- = N$), being the number of positive and negative samples, respectively.  In general, we may use shorthand $+$ and $-$, rather than the slightly more elaborate $+1$ and $-1$, e.g.\ $N_+ = N_{+1}$.

A classifier $C$ is a function or mapping from the space of feature vectors $\mathbb{R}^d$ to the set of labels $\{-1,+1\}$.
The input space on which $C$ is defined can be taken smaller than $\mathbb{R}^d$, for instance, if we know that some measurements can only take on continuous values between 0 and 1 or if they can only be integer.  At the least, the input space is typically larger than the set of $N$ training points as the aim is for our classifier to generalize to new and unseen input vectors. Here we generally consider all of $\mathbb{R}^d$ as possible inputs.
One way or the other, a classifier $C$ splits up the space in two sets.  One in which points are assigned to $+1$ and one in which the class assignment is $-1$.  The boundary between these two sets is generally referred to as the classification or decision boundary of $C$.

Finally, we define the objective that every classifier sets out to optimize.  In a sense, this will also act as the ultimate measure to decide which of two or more classifiers performs the best.  The de facto standard is the measure  $\varepsilon$ that determines the fraction of misclassified objects when applying classifier $C$ to the problem described by $p_{XY}$\footnote{This is not to say that other measures have not been studied.  Depending on the actual goal, measures like AUC, $F$ score, or the $H$ measure have been considered as well.  For both general and critical coverage of such measures refer to \cite{bradley1997use, fawcett2006introduction, hand2014better, hand2017note, lavravc1999rule, landgrebe2006precision, provost1997case}.\label{auc}}. Using Iverson brackets, we have
\begin{equation}\label{eq:err}
\varepsilon = \int_{\mathbb{R}^d} \sum_{y \in \{-1,+1\} } [ C(x) \neq y ] p_{XY}(x,y) dx \, .
\end{equation}
This measure goes under various names like (true) error rate, classification error, 0-1 risk, and probability of misclassification.  Then again, some may rather refer to it in terms of accuracy, which equals one minus the error rate.  The lower $\varepsilon$ for a particular classifier the better we would say it is for the particular problem $p_{XY}$, as it will perform better in expectation\footnote{Expanding a bit further on footnote \ref{auc}, we remark that one of the more important settings, in which another performance measure or another way of evaluating may be appropriate, is in the case where the classification cost per class is different.  Equation \eqref{eq:err} tacitly assumes that predicting the wrong class incurs a cost of one, while predicting the right class comes at no cost.  In many real-world settings, however, making the one error is not as costly as making the other.  For instance, building a rotten fruit detector, classifying a fresh piece of fruit as rotten could turn out less costly than classifying a bad piece of fruit as good.  When building an actual classifier, life often is even worse as one may not even know what the cost really is that will be incurred by a misclassification.  This is one reason to resort to an analysis of the so-called receiver operating characteristic (ROC) curve and its related measure: the area under the ROC curve (AUC, an abbreviation mention already in the previous footnote).  This curve and the related area provide, in some sense, tools to study the behavior of classifiers over all possible misclassification costs simultaneously.

Another important classification setting is the one in which there is a strong imbalance in class sizes, e.g. where we expect the one class to occur significantly much more often than the other class---a situation easily imagined in various applications.  Also here analyses through ROC, AUC, and related techniques are advisable.  For more on this topic, the reader is kindly referred to \cite{fawcett2006introduction}, \cite{hand2014better}, and related work.}.

\subsection{The Bayes Classifier}\label{sect:bayes}

One should understand that if we know $p_{XY}$, we are done.  In that case, we can construct an optimal classifier $C$ that attains the minimum risk $\varepsilon^\ast$.  But how do we get to that classifier?  Let $C^\ast$ refer to this optimal classifier, fix the feature vector that we want to label to $x$, and consider the corresponding term within the integral of Equation \eqref{eq:err}:
\begin{equation}
\sum_{y \in \{-1,+1\} } [ C^\ast(x) \neq y ] p_{XY}(x,y) = [ C^\ast(x) \neq +1 ] p_{XY}(x,+1) + [ C^\ast(x) \neq -1 ] p_{XY}(x,-1) \, .
\end{equation}
As $C^\ast$ should assign $x$ to $+1$ or $-1$, we see that the the choice that adds the least to the integral at this feature vector value is the assignment to that class for which $p_{XY}$ is largest.  We reach an overall minimum if we stick to this optimal choice in every location\footnote{Of course, if we wish, we generally can decide otherwise on a set of measure 0 without doing any harm to the optimality of the classifier $C^\ast$.} $x \in \mathbb{R}^d$.  In case $p_{XY}(x,+1)=p_{XY}(x,-1)$, where $x$ is actually on the decision boundary, it does not matter what decision the classifier makes, as it will induce an equally large error. In other words, we can define $C^\ast: \mathbb{R}^d \to \{-1,+1\}$ as follows:
\begin{equation}\label{eq:bayesc}
C^\ast(x) =
\begin{cases}
-1 & \mbox{if } p_{XY}(x,-1) > p_{XY}(x,+1) \\
+1 & \mbox{otherwise}
\end{cases} \, .
\end{equation}
Again using Iverson brackets, we could equally well write this as $C^\ast(x) = 1 - 2[ p_{XY}(x,-1) > p_{XY}(x,+1) ]$.

A possibly more instructive reformulation is by considering the conditional probabilities $p_{Y|X}$, often referred to as the posterior probabilities or simply the posteriors, instead of the full probabilities. Equivalent to checking $p_{XY}(x,-1) > p_{XY}(x,+1)$, we can verify whether $p_{Y|X}(-1|x) > p_{Y|X}(+1|x)$ and in the same vein as in Equation \eqref{eq:bayesc} decide to assign $x$ to $-1$ if this is indeed the case and assign it to $+1$ otherwise.  The latter basically states that, given the observations made, one should assign the corresponding object to the class with the largest probability conditioned on those observations.  Especially formulated like this, it seems like the obviously optimal assignment strategy.

The theoretical constructs $\varepsilon^\ast$ and $C^\ast$ are referred to as the Bayes error rate and the Bayes classifier, respectively.  The former gives a lower bound on the best error we could ever achieve on the problem at hand.  The latter shows us how to make optimal decisions once $p_{XY}$ is known.  But these quantities are merely of theoretical importance indeed.  In reality, our only hope is to approximate them, as the exact $p_{XY}$ will never be available to us.  The objects we can work with are the $N$ draws $(x_i,y_i)$ from that same distribution.  Based on these examples, we aim to build a classifier that generalizes well to all of $p_{XY}$.  In all that follows in this chapter, this is the setting considered.

\subsection{Generative Probabilistic Classifiers}

The previous subsection showed that if we have $p_{XY}$, we can compare $p_{XY}(x,-1)$ and $p_{XY}(x,+1)$, and perform an optimal class assignment.  One way to get a hold on $p_{XY}$ is by making assumptions on the form of the underlying class distributions $p_{X|Y}$ and estimate its free parameters from the training data provided.  Such class-conditional probabilities describe the distribution of the underlying individual classes, i.e., they consider the distribution of $X$ given $Y$.  Subsequently, these $p_{X|Y}$ can be combined with an estimate of the prior probability $p_Y$ of every class---i.e., an estimate of how often every class anyway occurs---to come to an overall estimate of $p_Y p_{X|Y} = p_{XY}$.  These models are called generative, because, once they are fitted to the data, they allow us to generate new data from these class-conditional distributions, though the accuracy of this generative process of course heavily depends on the accuracy of the model fit.

A very common, and one of the more simple instantiations of a generative model, is classical linear discriminant analysis (LDA)\footnote{Some authors take the term linear discriminant analysis to refer to a more broadly defined class of classifiers.  The normality-based classifier discussed here is also referred to by some as Fisher's linear discriminant or classical discriminant analysis.}, which assumes the classes to be normally distributed with different means and equal covariance matrices.  Denoting the normal distribution with mean $\mu$ and covariance $\Sigma$ by $g(\cdot|\mu,\Sigma)$, the full probability model can be written as
\begin{equation}\label{eq:lda}
p(x,y|\mu_{+1},\mu_{-1},\Sigma) =
\begin{cases}
\pi_+ g(x|\mu_+,\Sigma) & \mbox{if } y = +1 \\
\pi_- g(x|\mu_-,\Sigma) & \mbox{if } y = -1
\end{cases} \, ,
\end{equation}
where $\pi_+$ and $\pi_- = 1 - \pi_+$ are the class priors, which are in the $[0,1]$ interval.

The above merely specifies the class of models that we consider, but it does not tell us how we fit it to the data that we have.  A classical way to come to such parameters is to determine the maximum likelihood estimates. Other approaches, like maximum a posteriori and proper Bayesian estimation, are possible as well.  Relying on the log-likelihood, the objective function we find equals
\begin{equation}\label{eq:ll}
\begin{split}
L(\pi_+,\pi_-,\mu_+,\mu_-,\Sigma) = & \sum_{i=1}^N \log  p(x_i,y_i|\mu_+,\mu_-,\Sigma)  \\
 = & \sum_{i:y_i=+1} \left( \log \pi_+ + \log g(x_i|\mu_+,\Sigma) \right) \\
 + & \sum_{i:y_i=-1} \left( \log \pi_- + \log g(x_i|\mu_-,\Sigma) \right) \, .
\end{split}
\end{equation}
Maximizing this leads to the well-known closed-form estimators
\begin{align}
\hat{\pi}_+ &= \frac{N_+}{N} \\
\hat{\pi}_- &= \frac{N_-}{N} \\
\hat{\mu}_+ &= \frac{1}{N_+}\sum_{i:y_i=+1} x_i \\
\hat{\mu}_- &= \frac{1}{N_-}\sum_{i:y_i=-1} x_i \\
\hat{\Sigma} &= \frac{1}{N}\sum_{i=1}^N (x_i - \hat{\mu}_{y_i})(x_i - \hat{\mu}_{y_i})^\top \, .
\end{align}
We note that, no matter what estimates one chooses, the L in the abbreviation LDA refers to the fact that the decision boundary forms a $(d-1)$-dimensional hyperplane, as can be checked by explicitly solving $\pi_+ g(x|\mu_+,\Sigma) = \pi_- g(x|\mu_-,\Sigma)$ for $x$ and finding that it takes on the form of a linear equation\footnote{It actually is a nonhomogeneous linear or affine one, to be a bit more precise.}:
\begin{equation}\label{eq:ldaboundary}
(\hat{\mu}_+ - \hat{\mu}_-)^\top \hat{\Sigma}^{-1} x + c = 0\, ,
\end{equation}
with $c$ a constant offset that depends on the dimensionality $d$, the determinant of $\hat{\Sigma}$, and the prior probabilities $\hat{\pi}_+$ and $\hat{\pi}_-$.

Again, these estimates are just one way to specify the free parameters in the model.  For instance, $\Sigma$ could also have been estimated by an estimator such as $\frac{N}{N-2}\hat{\Sigma}$ or we could have applied Laplace smoothing to our prior estimates and, for instance, consider $(N_++1)/(N+2)$ instead of $N_+/N$.  Of course, probably of a more dramatic influence is the actual choice of normal class-conditional models, which just may not be realistic. Even though having a misspecified model, as such, does not mean that the classifier will not perform well, more advanced and complex choices for these class-conditional distributions have been introduced that can potentially improve upon the use of normal distributions (see, for instance, \cite{bishop1995neural, hastie2001elements, mclachlan2004discriminant, rasmussen2006gaussian, ripley2007pattern}, but let us also refer already to Subsection \ref{sect:singlebest} for some important notes regarding classifier complexity in relation to training set size).  As an example, we may substitute our relatively rigid parametric choice for a highly flexible nonparametric kernel density estimator per class, which leads to a classifier that is often referred to as the Parzen classifier.

\subsection{Discriminative Probabilistic Classifiers}\label{sect:discr}

In the previous subsection, we decided to model $p_{XY}$, based on which we can then come to a decision on whether to assign $o_i$ to the $+$ or the $-$ class considering its corresponding feature vector $x_i$.  Subsection \ref{sect:bayes}, however, showed that we might as well use a model of $p_{Y|X}$ to reach a decision.  Of course, from the full model $p_{XY}$, we can get to the conditional $p_{Y|X}$, while going into the other direction is not possible.  For classification, however, we merely need to know $p_{Y|X}$ and so we can save the trouble of building a full model.  In fact, if we are unsure about the true form of the underlying class-conditionals or the marginal $p_X$ that describes the feature vector distribution, directly modeling $p_{Y|X}$ may be wise, as we can avoid potential problems due to such model misspecification.  On the other hand, if the full model is accurate enough this may have a positive effect on the classifier's performance \cite{mclachlan2004discriminant, rubinstein1997discriminative}.  Approaches that directly model $p_{Y|X}$ are called discriminative as they aim to get straightaway to the information that matters to tell the one class apart from the other.

The classical model in this setting, and in a sense a counterpart of LDA, is called logistic regression\footnote{Again, there are various other names that refer to, at least more or less, the same approach, e.g.\ the logistic classifier, logistic discrimination, or logistic discriminant analysis.}.  One way to get to this model is to assume that the logarithm of the so-called posterior odds ratio takes on a linear form in $x$, i.e.,
\begin{equation}\label{eq:logodds}
\log \frac{p_{Y|X}(+1|x)}{p_{Y|X}(-1|x)} = w^\top x + w_\circ \, ,
\end{equation}
with $w \in \mathbb{R}^d$ and $w_\circ \in \mathbb{R}$.  From this we derive that the posterior for the positive class takes on the following form:
\begin{equation}\label{eq:sigmoid}
p_{Y|X}(+1|x) = 1- p_{Y|X}(-1|x) = \frac{\exp(w^\top x + w_\circ)}{1 + \exp(w^\top x + w_\circ)} = 1 - \frac{1}{1 + \exp(w^\top x + w_\circ)} \, .
\end{equation}
The parameters $w$ and $w_\circ$ again are typically estimated by maximizing the log-likelihood.  Formally, we have to consider the likelihood of the full model and not only of its posterior, but the choice of the necessary additional marginal model for $p_X$ is of no influence on the optimum of the parameters we are interested in \cite{mclachlan2004discriminant, minka2005discriminative} and so we may just consider
\begin{equation}\label{eq:logisticml}
(\hat{w},\hat{w}_\circ) = \operatorname*{argmax}_{(w,w_\circ) \in \mathbb{R}^{d+1}} \sum_{i=1}^N \log \left( \frac{y_i+1}{2} -  \frac{y_i}{1 + \exp(w^\top x_i + w_\circ)} \right) \, .
\end{equation}
Note that, like LDA, this classifier is linear as well.  Generally, the decision boundary is located at the $x$ for which $p_{XY}(x,+1)=p_{XY}(x,-1)$ or, similarly, for which $p_{Y|X}(+1|x)=p_{Y|X}(-1|x)$.  But the latter case implies that the log-odds equals $0$ and so the decision boundary takes on the form
\begin{equation}\label{eq:logisticboundary}
\hat{w}^\top x_i + \hat{w}_\circ = 0 \, .
\end{equation}

As for generative models, discriminative probabilistic ones come in all different kinds of flavors.  A particularly popular and fairly general form of turning linear classifiers into nonlinear ones is discussed in Subsection \ref{sect:feattrans}.  These and more variations can be found, among others, in \cite{bishop1995neural, hastie2001elements, hinton1989connectionist, mclachlan2004discriminant, rasmussen2006gaussian, ripley2007pattern}.

\subsection{Losses and Hypothesis Spaces}\label{sect:landh}

As made explicit in Equations \eqref{eq:ldaboundary} and \eqref{eq:logisticboundary}, both LDA and logistic regression lead to linear decision boundaries, i.e., $(d-1)$-dimensional hyperplanes in a $d$-dimensional feature space.  The functional form of these decision boundaries is indirectly fixed through the choice of probabilistic model.  Ultimately, they can be seen as mapping a feature vector $x$ in a linear way to a value on the real line. A subsequent decision to assign this point to the $+$ or $-$ class depends on whether the obtained value is smaller or larger than $0$.  This last operation, which basically turns the linear mapping into an actual classifier, can in essence be carried out by applying the $\sign$ function following the linear transform. This maps every number to a point in the set $\{-1,+1\}$, except for the points on the decision boundary which are mapped to $0$\footnote{As this is typically a set of measure zero, where the decision boundary gets mapped to does not really influence the performance of the classifier.  Nevertheless, one may of course decide to have the boundary points mapped to $-1$ or $+1$, rather than $0$.}.  The main difference between LDA and logistic regression is then the way the free parameters of their respective linear transformations have been obtained.

Especially within some areas of machine learning, it is common to altogether forget about the potentially probabilistic interpretation of a classifier. Instead, one explicitly defines the set $H$ of functions that transform any feature vector into a single number and which can subsequently be turned into actual classifiers by ``signing'' them.  Next to that, one defines a measure $\ell:\mathbb{R} \times \{-1,+1\} \to \mathbb{R}$ of how good or bad a particular function fits to any point in the training data.  The set of functions is often referred to as the hypothesis space, while the measure of fit is typically referred to as the loss (which is something that one minimizes).  More specifically, the loss function $\ell$ takes in a predicted value $h(x)$, with $h \in H$, and decides how good this value matches the desired output $y$.  Once, $H$ and $\ell$ are fixed, we can search for a best fitting $h^* \in H$,
\begin{equation}\label{eq:opthyp}
h^* = \operatorname*{argmin}_{h \in H} \sum_{i=1}^N   \ell(h(x_i),y_i) \, ,
\end{equation}
which can then be used to classify new and unseen object $x$ by assigning it to class $\sign(h^*(x))$.

The foregoing is a fairly general way of formulating the training of a classifier and not every choice for $H$ and $\ell$ may be equally convenient or suitable to be used on a classification problem.  In the following, we present some of the better-known choices and briefly introduce some of the classifiers related to the particular options.

\subsubsection{0-1 Loss}

In a way, the obvious choice is to take the loss that we are actually interested in: the fraction of misclassified observations.  Equation \eqref{eq:err} defines this fraction, i.e., the classification error or the 0-1 risk, under the true distribution.  Considering our finite number $N$ of training data, the best we can do is just count the number of incorrectly assigned samples:
\begin{equation}\label{eq:0-1risk}
L_\text{0-1}(h) := \sum_{i=1}^N \ell_\text{0-1}(h(x_i),y_i) \, ,
\end{equation}
with
\begin{equation}\label{eq:0-1loss}
\ell_\text{0-1}(a,b) := [\sign(a) \neq b]
\end{equation}
being the 0-1 loss\footnote{We note here that the expected loss is often referred to as the risk, which explains why the error defined in Equation \eqref{eq:err} is also referred to as the 0-1 risk.}.

A major problem with this loss is that finding the optimal $h^*$ is for many cases computationally very hard.  Take for $H$, for example, all linear functions, then finding our $h^*$ turns out to be NP-hard and even settling for approximate solutions does not necessarily help \cite{ben2003difficulty, hoffgen1995robust}.

\subsubsection{Convex Surrogate Losses}

The most broadly used approach to get around the problem of the computational complexity that the 0-1 loss poses is to consider a relaxation of the original problem, which turns it into a convex optimization problem that is relatively easy to solve, e.g. by gradient descent or variations of this technique.

To achieve this, first of all, one chooses $H$ to be convex.  The classical choice would be the space of linear functions, but more complex choices are possible as we will see later.  The second step in the relaxation of the optimization problem is to turn to so-called surrogate losses.  These are losses that aim to approximate $\ell_\text{0-1}$ as well as possible, but have some additional benefits.  For instance, one could choose a loss that is everywhere differentiable, something that does not hold for the 0-1 loss.  To turn the optimization for $h$ into a convex optimization, the function in Equation \eqref{eq:0-1risk} needs to be convex as well.  To generally do so, we approximate $\ell_\text{0-1}$ in Equation \eqref{eq:0-1loss} by a convex function that upper-bounds this loss everywhere.  The idea of the upper bound is that if we find the minimizer to it, we know that 0-1 loss one would achieve on the training set is certainly not larger than the surrogate risk the minimizer attains.

\subsubsection{Particular Surrogate Losses}

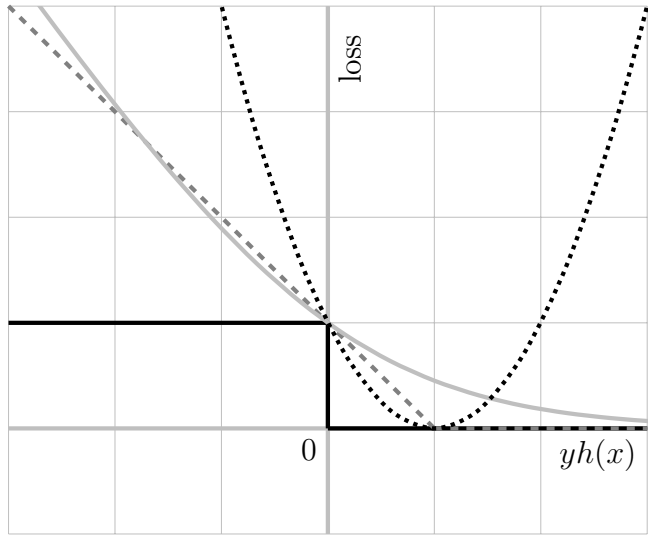
\begin{figure}
\center
\begin{tikzpicture}[scale=1.4]
\draw[step=1cm,lightgray,very thin] (-3,-1) grid (3,4);
\draw[ultra thick,lightgray,-] (-3,0) -- (3,0);
\draw[black](3,0) node[anchor=north east] {$yh(x)$};
\draw[ultra thick,lightgray,-] (0,-1) -- (0,4);
\draw[black](0,3.5) node[anchor=north, rotate=90] {loss};
\draw (0,0) node[anchor=north east] {0};
\draw[ultra thick,black,-] (-3,1) -- (0,1);
\draw[ultra thick,black,-] (0,1) -- (0,0);
\draw[ultra thick,black,-] (0,0) -- (3,0);
\draw[ultra thick,gray,dashed] (-3,4) -- (1,0);
\draw[ultra thick,gray,dashed] (1,0) -- (3,0);
\clip (-3,-2) rectangle (3,4);
\draw[ultra thick,lightgray,domain=-4:6,smooth,variable=\x] plot ({\x},{ln(1+exp(-\x))/ln(2)});
\clip (-3,-2) rectangle (3,4);
\draw[ultra thick,dotted,domain=-4:6,smooth,variable=\x] plot ({\x},{(1-\x)^2});
\end{tikzpicture}
\caption{Plot of the 0-1 loss in terms $y$ times the predicted output $h(x)$ (in solid black) and three examples of surrogate losses that upperbound the 0-1 loss.  Logistic loss is in solid light gray, hinge loss is in dashed dark gray, and squared loss is dotted black.}\label{fig:loss}
\end{figure}

As we would decide on the label of a sample $x$ based on the output $h(x)$ that a trained classifier $h \in H$ provides, one typically only needs to consider what the loss function does for the value $yh(x)$.  For instance, we can rewrite $\ell_\text{0-1}(a,b)$ as $\ell_\text{0-1}(a,b) = \ell_\text{0-1}(ba) = [b\,\sign(a) \neq 1]$ and achieve the same loss.  In Figure \ref{fig:loss} the shape of the 0-1 loss is plotted in these terms.  The same figure shows various widely-used upper bounds for $\ell_\text{0-1}$.

Maybe the first one to note is the logistic loss, which is defined as
\begin{equation}\label{eq:logloss}
\ell_\text{logistic}(a,b) := \log_2 \left( 1 + e^{-ba} \right) \, .
\end{equation}
The figure displays it as the solid light gray curve.  Using this loss, in combination with a linear hypothesis class, leads to standard logistic regression as introduced in Subsection \ref{sect:discr}.  So in this case, we have both a probabilistic view of the resulting classifier as well as an interpretation of logistic regression as minimizer of a specific surrogate loss.  A second well-known classifier, or at least a basic form of it, is obtained by using the so-called hinge loss:
\begin{equation}\label{eq:hingeloss}
\ell_\text{hinge}(a,b) := \max(1 - ba,0) \, .
\end{equation}
This loss is at the basis of the support vector machine\footnote{Together with the underlying theory, SVMs caused all the furore in the late 1990s and early 2000s.  To many, the development of the SVM may still be one of the prime achievements of the mathematical field of statistical learning theory that started with Vapnik and Chervonenkis in the early 1970s.  At least, SVMs are still one of the most widely known and used classifiers within the fields of pattern recognition and machine learning.  Possibly one of the main feats of statistical learning theory was that it broke with the statistical tradition of merely studying the theory of the asymptotic behavior of estimators. Statistical learning theory is also concerned with the finite sample setting and makes, for instance, statements on the expected performance on unseen data for classifiers that have been trained on a limited set of examples \cite{christianini2000support, devroye1996probabilistic, vapnik1998statistical}.} (SVM) \cite{boser1992training, christianini2000support, cortes1995support, vapnik1998statistical}.   A third classifier that fits the general formalism and is widely employed is obtained by using the squared loss function
\begin{equation}\label{eq:lsloss}
\ell_\text{squared}(a,b) := (1 - ba)^2 \, .
\end{equation}
Using again the set of linear hypotheses, we get basically what is, among others, referred to as the linear regression classifier, the least squares classifier, the least squares support vector machine, the Fisher classifier, or Fisher's linear discriminant \cite{duda1973pattern, hastie2001elements, poggio2003mathematics, suykens1999least}.  Indeed, this classifier is a reinterpretation of the classical decision function introduced by Fisher \cite{fisher1936use} in the language of losses and hypotheses.

Finally, other losses one may encounter in the literature are the exponential loss $\exp(-ba)$, the truncated squared loss $\max(1 - ba,0)^2$, and the absolute loss $|1 - ba|$.  In Subsection \ref{sect:feattrans}, we introduce ways of designing nonlinear classifiers, which often rely on the same formalism as presented in this subsection.

\subsection{Neural Networks}

The use of artificial neural networks for supervised learning can be traced back at least to 1958.  In that year, the perceptron was introduced \cite{rosenblatt1958perceptron}, providing a linear classifier that could be trained using a basic iterative updating scheme for its parameters.  Currently, neural networks are the dominant technique in many applications and application related areas when massively sized data sets to train from are available.

Even though the original formulation of the perceptron does not give a direct interpretation in terms of the solution to the optimizing of a particular loss within its hypothesis space of linear classifiers, such formulations are possible \cite{devroye1996probabilistic}.  Neural networks that are employed nowadays readily follow the earlier sketched loss-hypothesis space paradigm.   Possibly the most characteristic feature of neural networks is that the hypotheses considered are built up of relatively simple computational units called neurons or nodes.  Such unit is a function $g:\mathbb{R}^q \to \mathbb{R}$ that takes in $q$ inputs and maps these to a single numerical output.  Typically, $g$ takes on the form of a linear mapping followed by a nonlinear transfer or activation function $\sigma:\mathbb{R} \to \mathbb{R}$:
\begin{equation}\label{eq:node}
  g(x) = \sigma (w^\top x + w_\circ) \, .
\end{equation}
Often $\sigma$ is taken to be of sigmoidal shape, like the logistic function in Equation \eqref{eq:sigmoid} used in logistic regression.  This function squeezes the real line onto the interval $[0,1]$, resembling a smooth threshold function.  Other choices are possible however.  A choice popularized more recently, with a clearly different characteristic is the so-called rectified linear unit, which is defined as $\sigma(x)=\max(0,x)$ \cite{nair2010rectified}.  As for various of the previously mentioned classifiers, the free parameters are tuned by measuring how well $g$ fits the given training data.  A widely used choice is the squared loss, but also likelihood based methods have been considered and links with probabilistic models have been studied \cite{bishop1995neural, hinton1989connectionist, ripley2007pattern}.

One should realize that, whatever the choice of activation function, as long as it is monotonic using $g$ for classification will lead to a linear classifier.  Nonlinear classifiers are constructed by combining various $g$s, both in parallel and in sequence. In this way, one can build arbitrarily large networks that can perform arbitrarily complex input-output mappings.  This means that we are dealing with large and diverse hypothesis classes $H$.  The general construction is that multiple nodes, connected in parallel, provide the inputs to subsequent nodes.  Consider, for instance, the nonlinear extension where, instead of a single node $g$, as a first step, we have multiple nodes $g_1, \ldots, g_D$ that all receive the same feature vectors $x$ as input.  In a second step, these $D$ outputs are collected by yet another node, $g:\mathbb{R}^D \to \mathbb{R}$ and transformed in a similar kind of way.  So, all in all, we get a function $G$ of the form
\begin{equation}\label{eq:multinode}
  G(x) = g\left(g_1(x),\ldots,g_D(x)\right) = \sigma \left(\sum_{i=1}^D w_i \sigma(w_{ij}^\top x + w_{i\circ}) + w_\circ \right) \, .
\end{equation}
To fully specify a particular $G$, one needs to set all the parameters in all $D+1$ nodes.  Once these are set, we can again use it to classify any $x$ to the sign of $G(x)$.

Of course, one does not have to stop at two steps.  The network can have an arbitrary number of steps, or layers as they are typically referred to.  Nowadays, so called deep\footnote{There is, of course, not clear-cut definition at what number of layers makes a network deep. This is similar to wondering what amount of data makes it big or what number of features makes a problem high-dimensional.} networks are being employed with hundreds of layers and millions of parameters.  In addition to this, there are generally many different variations to the basic scheme we have sketched here \cite{schmidhuber2015deep}.  By making different choices for the transfer function, by using multiple transfer functions, by changing the structure of the network, the number of nodes per layer, etc., one basically changes the hypothesis class that is considered.  In addition, where in Subsection \ref{sect:landh} the choice of $H$ and $\ell$ would typically be such that we end up with a convex optimization problem, using neural networks, we typically move away from optimization problems for which one can reasonably expect to be able to find the global optimum.  As a result, to fully define the classifier, we should not only specify the loss and the hypothesis class, but also the exact optimization procedure that is employed.  There are many choices possible to carry out the optimization, but most approaches rely on gradient descent or variations to this basic scheme \cite{bishop1995neural, bottou91c, hinton1989connectionist, white1989learning}.

\subsection{Neighbors, Trees, Ensembles, and All That}

There are a few approaches to classification that do not really fit the aforementioned settings, but that we feel do need brief mentioning either for historical reasons or for completeness.

\subsubsection[$k$ Nearest Neighbors]{$\bm{k}$ Nearest Neighbors}

The nearest neighbor rule \cite{fix1951discriminatory, cover1967nearest} is maybe the classifier with the most intuitive appeal.  It is a widely used and classical decision rule and one of the earliest nonparametric classifiers proposed.  In order to classify a new and unseen object, one simply determines the distances between its describing feature vector and the feature vectors in the training set and decides to assign the object to the same class the closest feature vector in that training set has.  Most often, the Euclidean distance is used to determine the nearest neighbor in the training data set, but in principle any other, possibly even expert-designed or learned, distance measure can be employed.

A direct, yet worthwhile extension is to not only consider the closest sample in the training set, i.e., the first nearest neighbor, but to consider the closest $k$ and assign any unseen object to the class that occurs most often among these $k$ nearest neighbors.  The $k$ nearest neighbor classifier, has various nice and interesting properties \cite{fix1951discriminatory, cover1967nearest, devroye1996probabilistic}. One of the more interesting ones may be the result that roughly states that with increasing numbers of training data, the $k$ nearest neighbor classifier converges to the Bayes classifier $C^*$, given $k$ increases at the appropriate rate.

\subsubsection{Decision Trees}

Classification trees or decision trees are classifiers that can be visualized by a hierarchical tree structure \cite{breiman1984classification, quinlan1986induction}.  Every observation that is classified traverses the tree's nodes to arrive at a leave that contains the final decision, i.e., assign label $+1$ or $-1$.  In every node, a basic operation decides what next node, at a level deeper, will be visited.

Probably the simplest and most extensively used form of decisions that are made at every node are those that just check whether or not a single particular feature value is larger than a chosen threshold value.  As there are only two outcomes possible at every step, there will only be two nodes available at every next level that can be reached.  Using training data, there are various ways to come to an automated construction of the exact tree hierarchy, together with the decisions to make at the different nodes.

One of the possible benefits of these kinds of classifiers, especially of the last mentioned type of decision tree, given the tree is not too deep, is that one can easily trace back and interpret how the decision to assign a sample to class $-1$ or $+1$ was reached.  One can retrace the steps through the tree and see what subsequent decisions lead to the class label provided.

\subsubsection{Multiple Classifier Systems}\label{sect:mcs}

The terms multiple classifier systems, classifier combining, and ensemble methods all refer to roughly the same idea: potentially more powerful classifiers can be built by combining two or more of them \cite{kittler2000multiple, kuncheva2004combining, polikar2006ensemble}.  The latter are often referred to as base\footnote{That is, \textbackslash\textprimstress b\={a}s\textbackslash rather than \textbackslash\textprimstress b\={a}z-\textbackslash .} classifiers.  So, these techniques are not classifiers as such, but ways to compile base classifiers into classifiers that in some sense befit the data better.  There can be various reasons to combine classifiers.

Sometimes a classifier turns out to be overly flexible and one may wish to stabilize the base classifier (see also Section \ref{sect:reg}).  One way to do so is by a well-known combining technique called bagging \cite{breiman1996bagging}, which trains various classifiers based on bootstrap samples of the same data set and assigns any new sample based on the average output of this often large set of base classifiers.  Another way to construct different base classifiers is to consider random subspaces by sampling a set of different features for every base learner.  This technique has been extensively exploited in random forests and the like \cite{ho1995random ,ho1998random}.

Combining classifiers can also be exploited when dealing with a problem where, in some sense, essentially different sets of features play a role.  For instance, in the analysis of patient data, one might want to use different classifiers for high-dimensional genetic measurements and low-dimensional clinical data, as these sets may behave rather differently from each other.  Once the two or more specialized classifiers have been trained, various forms of so-called fixed and trained combiners can be applied to come to a final decision rule \cite{duin2002combining, kuncheva2004combining}.

At times, the base classifiers can already be quite complex, possibly being a multiple classifier system in itself.  Nice examples are available from medical image analysis \cite{niemeijer2011combining} and recommender systems \cite{jahrer2010combining}\footnote{Though illustrative, strictly speaking, this work does not report on a classification task.}.  In these cases, advanced systems have been developed independently from each other.  As a result, there is a fair chance that every systems has its own strengths and weaknesses and even the best performing individual system cannot be expected to perform the best in every part of feature space. Hence combining such systems can result in significantly improved overall performance.

Another reason to employ classifier combining is to integrate contextual features into the classification process.  Such approaches can especially be beneficial when integrating contextual information into image and signal analysis tasks \cite{cohen2005stacked, loog2002supervised, loog2004supervised}.  These techniques can be seen as a specific form of stacked generalization or stacking \cite{wolpert1992stacked} and are becoming relevant again these days in the connection with deep learning (see, for instance, \cite{fu2016occlusion, li2016iterative, shrivastava2016contextual}).

Finally, we should mention boosting approaches to multiple classifier systems and in particular adaboost \cite{freund1995desicion}.  Boosting was initially studied in a more theoretical setting to show that so-called weak learners, i.e., classifiers that barely reach better performance than an error rate equal to the a priori probability of the largest class, could be combined into a strong learner to significantly improve performance over the weak ones.  This research culminated in the development of a combining technique that sequentially adds base classifiers to the ensemble that has already been constructed, where the next base classifier focuses especially on samples that previous base learners were unable to correctly classify. This last feature is the adaptive characteristic of this particular combining scheme that warrants the prefix ada-.

\section{Representations and Classifier Complexity}\label{sect:featrep}

In all of the foregoing, we tacitly assumed that we already had decided on what feature set to use for every object or, more generally, on how to represent every object before constructing our classification rule.  In actual applications, there are, however, choices to be made.

To start with, we may be lucky and even have influence on what kind of raw measurements are going to be taken.  Are we making color images with a standard camera of our oranges?  Do we make a CT or an MR scan?  Do we measure weight, diameter, plasticity?  Will we sequence tissue samples?  Clearly, the actual measurements that are most informative depend on the classification task at hand.  More importantly, one should realize that once key measurements have been omitted\footnote{Such situation may of course arise from the fact that particular measurements simply cannot be realized for a number of reasons.  For instance, measurements can be too expensive to consider as part of a realistic solution or they can be invasive to an unacceptable degree.}, one can never recover the lost information through the building of a classifier, no matter how advanced or clever the classifier one designs is.

In reality, the practitioner often has little influence on what precise measurements are going to be made and s/he has to work with a predetermined set of initial features.  But even in this setting there are considerations to be made.  Are we going to use all of these features?  Are we using them as is, or do we construct derivative features.  Such decisions can hinge, say, on one's own insight into the problem or on possible external expert knowledge that one has acquired.  Do we really need the RGB values of every pixel when describing our oranges or is an average value per channel enough?  Do we at all need RGB or is a measure of orangeness sufficient?  And do we need that weight measurement or does our consulting expert suggest that it does not contain any relevant information and can we do without?

We give a brief overview of ways to represent objects and some tools with which we can partially automate the process of selecting features and create derivatives. In addition, we discuss some of the---initially maybe counterintuitive---effects of using more and more measurements and how this roughly relates to the complexity of the classifier. We also introduce a tool possibly valuable in the analysis of the effect of the number of features: the so-called feature curve.

\subsection{Feature Transformations}\label{sect:feattrans}

In the previous section, we saw various linear classifiers like logistic regression, SVMs, and Fisher's discriminant.  At first sight, the linearity of a classifier may seem like a limitation, but this is actually easily removed.  Nonlinear classifiers can be created by including nonlinear transformations of the original features into the original feature vector and subsequently train a linear classifier in this extended space.  This essentially extends the hypothesis space $H$.  For example, if $x_i \in \mathbb{R}^2$, then we can form new and nonlinear features by transforming the original individual features, e.g.\ we can transform the first feature $x_{i1}$ into $x_{i1}^3$ or $\sin x_{i1}$, and by combining individual features, e.g.\
we can form the product feature $x_{i1} x_{i2}$.  Clearly, the possibilities are endless.

Let $\varphi:\mathbb{R}^d \to \mathbb{R}^D$ be the transformation that maps the original feature vector to its extended representation.  Once we learned a classifier $C_D$ in the extended space, it induces a classifier in the original space through $C_d = C_D \circ \varphi$: we simply take every feature vector to be classified, transform it by $\varphi$, and apply the classifier trained in the higher-dimensional space.  Using this construct, even if we would limit ourselves to linear classifiers in $\mathbb{R}^D$, we would typically find nonlinear decision boundaries in the original $d$-dimensional feature space.

\subsubsection{The Kernel Trick}\label{sect:kernel}

SVMs not only received a lot of attention as a result of statistical learning theory, the SVM literature also introduced what has become widely known as the kernel trick or kernel method \cite{boser1992training}, which has its roots in the 1960s \cite{aizerman1964theoretical}.  The kernel trick allows one to extend many inherently linear approaches to nonlinear ones in a computationally simple way.  At its basis is that---following the representer theorem \cite{scholkopf2001generalized, wahba1990spline}---many solutions for the type of optimization problems for linear classifiers that we have considered in Subsection \ref{sect:landh} can be expressed in terms of a weighted combination of inner products of training feature vectors and the $x$ that is being classified, i.e.,
\begin{equation}\label{eq:toptim}
  h^*(x) = w^\top x + w_\circ = \sum_{i=1}^N a_i x_i^\top x + a_\circ \, ,
\end{equation}
with $a_i \in \mathbb{R}$.  Therefore, finding $h^*$ becomes equivalent to finding the optimal coefficients $a_i$.

After mapping the original feature vectors with $\varphi$, we would be optimizing the equivalent in the $D$-dimensional space to get to a possibly nonlinear classifier:
\begin{equation}\label{eq:koptim}
  h^*(x) = w^\top x + w_\circ = \sum_{i=1}^N a_i \varphi(x_i)^\top \varphi(x) + a_\circ \, .
\end{equation}
It becomes apparent that the only thing that matters in these settings is that we know how to compute inner products $k(z,x) := \varphi(z)^\top \varphi(x)$ between any two mapped feature vectors $x$ and $z$.  The function $k$ is also referred to as a kernel function or simply a kernel.  Of course, once we have explicitly defined $\varphi$, we can always construct the corresponding kernel functions, but the power of the kernel trick is that in many settings this can be avoided.  This is interesting in at least two ways.

The first one is that if one wants to construct highly nonlinear classifiers, the explicit expansion $\varphi$ could grow inconveniently large.  Take a simple expansion in which we consider all (unique) second degree monomials, which number equals $\binom{d}{2}$. So the dimensionality $D$ of the feature space in which we have to take the inner product grows as $O(d^2)$.  By a direct calculation, one can however show that the inner product in this larger space can be expressed in terms of a much simpler $k$.  In this case particularly, we have that\footnote{This can be demonstrated by explicitly writing out both sides of the equation.}
\begin{equation}\label{eq:kinner}
  \varphi(z)^\top \varphi(x) = (z^\top x)^2 \, .
\end{equation}
As one can imagine, moving to nonlinearities of even higher degree, the effect becomes more pronounced\footnote{We note that first degree monomials can also be included, either by explicitly including an additional feature to the original feature vector that is constant, say $c$, or implicitly by defining the inner product as $(z^\top x + c^2)^2$.}.  At some point, explicitly expanding the feature vector nonlinearly becomes prohibitive, while calculating the induced inner product may still be easy to do.  An extreme example is the radial basis function or Gaussian kernel defined by
\begin{equation}\label{eq:gaussk}
  k(z,x) = \exp \left( -\frac{1}{\sigma^2} || x-z ||^2  \right) \, ,
\end{equation}
which corresponds to a mapping that takes the original $d$-dimensional space to an infinite dimensional expansion \cite{vapnik1998statistical}.

A second reason why the formulation in terms of inner products is of interest is that it, in principle, allows us to forget about an explicit feature representation altogether.  Going back to our original objects $o_i$, if we can construct a function $k(\cdot,\cdot) \mapsto \mathbb{R}_0^+$ that takes in two objects and fulfils all the criteria of an kernel, we can directly use $k(o_i,o)$ (with $o$ the object that we want to classify) as a substitute for $\varphi(x_i)^\top \varphi(x)$ in Equation \eqref{eq:koptim}.  Once such a kernel function $k$ has been constructed---whether it is through an explicit feature space or not, one can use it to build classifiers.

All in all, kernel methods define a very general, powerful, and flexible formalism, which allows the design of problem specific kernels. Research into this direction has spawned a massive amount of publications about such approaches (see, for instance, \cite{christianini2000support, scholkopf2002learning}).

\subsection{Dissimilarity Representation}\label{sect:dissim}

Any kernel $k$ provides, in a way, a similarity measure between two feature observations $x$ and $z$ (or possible directly between two objects): the larger the value is the more similar the two observations are.  As $k$ has to act like an inner product that, at least implicitly, corresponds to some underlying feature space, limitations apply.  In many settings, one might actually have an idea of a proper way to measure the similarity or the, in some sense equivalent, dissimilarity between two objects\footnote{Depending on the requirements one imposes upon dissimilarities (or, proximities, distances, etc.), similarities $s$ can be turned into dissimilarities $\delta$.  For instance, by taking $\delta = \tfrac{1}{s}$ or $\delta = -s$.  Next to these very basic transforms, there are various more advanced possibilities to construct such conversions \cite{pekalska2005dissimilarity}.}.  Possibly, such measure is provided by an expert that is working in the field where you are asked to build your classifier for. It therefore may be expected to be a well thought-through quantity that captures the essential resemblance of or difference between two objects.

The dissimilarity approach \cite{duin1997experiments, pekalska2001generalized, pekalska2005dissimilarity} allows one to build classifiers similar to kernel-based classifiers, but without some of the restrictions.  One of the core ideas is that every objects can be represented, not by what one can see as absolute measurements that can be performed on every individual object, but rather by relative measurements that tells us how (dis)similar the object of interest is with a set of $D$ representative objects.  These representative objects are also referred to as the prototypes.  In particular, having such a set of prototypes $p_i$ with $i \in \{1, \ldots D\}$, and having our favorite dissimilarity measure $\delta$, every object $o$ can be represented by the $D$-dimensional dissimilarity vector
\begin{equation}\label{eq:disvect}
  x = (\delta(p_1,o), \ldots, \delta(p_D,o))^\top \, .
\end{equation}
Training, for instance, a linear classifier in this space leads to a hypothesis of the form
\begin{equation}\label{eq:disopt}
  w^\top x + w_\circ = \sum_{i=1}^N a_i \delta(p_i,o) + w_\circ \, ,
\end{equation}
which should be compared to Equation \eqref{eq:koptim}.

The linear classifier is just one example of a classifier one can use in these $D$ dimensions. In this dissimilarity space, one can of course use the full range of classifiers that have been introduced in this chapter.

\subsection{Feature Curves and the Curse of Dimensionality}\label{sect:featcurv}

Measuring more and more features on every object seems to imply that we gather more and more useful information on them.  The worst that can happen is that we measure features that are partly or completely redundant, e.g.\ measuring the density, while we already have measured the mass and the volume.  But once we have the information present in the features, it cannot vanish anymore.  In a sense this is indeed true, but the question is whether we can still extract the information relevant with growing feature numbers.  All classifiers rely on some form of estimation, which is basically what we do when we train a classifier, but estimation typically becomes less and less reliable when the space in which we carry it out grows\footnote{One can say that, in this specific sense, the complexity of the classifier increases.}.  The net result of this is that, while we typically would get improved performance with every additional feature in the very beginning, this effect will gradually wear off, and in the long run even leads to a deterioration in performance as soon as the estimates become unreliable enough.  This behavior is what is often referred to as the curse of dimensionality \cite{bishop1995neural, duin2015, hastie2001elements, jain2000statistical}.

A curve that plots the performance of a classifier against an increasing number of features is called a feature curve \cite{duin2015}.  It can be used as a simple analytic tool to get an idea of how sensitive our classifier is to the number of measurements that each object is described with.  Possibly of equal importance is that such curves can be used to compare two or more classifiers with each other.  The forms feature curves take on depends heavily on the specific problem that we are dealing with, on the complexity of the classification method, the way this complexity relates to the specific problem, and on the number $N$ of training samples we have to train our classifier. As far as it is at all possible, the exact mathematical quantification of these quantities is a real challenge.  Very roughly, one can state that the more complex a classifier is, the quicker its performance starts deteriorating with an increasing number of features.  On the other hand: the more training data that is available, the later the deterioration in performance sets in.  Also, the one classification technique is more complex than the other if the possible decision boundaries the former can model are more flexible or, similarly, less smooth.  Another way to think about this is that the hypothesis of the former classification method is larger than the latter one\footnote{Vapnik was one of the first to concern himself with quantifying  classifier complexity in a mathematically rigorous fashion.  Details can be found in his book on learning theory: \cite{vapnik1998statistical}.}.

\subsection{Feature Extraction and Selection}

The curse of dimensionality indicates that in particular cases it can be beneficial for the performance of our classifier to lower the feature dimensionality.  This may be applicable, for instance, if one has little insight in the classification problem at hand, in which case one tends to define lots of potentially useful features and/or dissimilarities in the hope that at least some of them pick up what is important to discriminate between the two classes.  Carrying out a more or less systematic reduction of the dimensionality after defining such large class of features can lead to acceptable classification results.

Roughly speaking, there are two main approaches \cite{guyon2003introduction, hastie2001elements, jain1997feature, ripley2007pattern}.  The first one is feature selection and the second one is feature extraction.  The former reduces the dimensionality by picking a subset from the original feature set, while the latter allows the combination of two or more features into fewer new features.  This combination is often restricted to linear transformations, i.e., weighted sums, of original features, meaning that one considers linear subspaces of the original feature space. In principle, however, feature extraction also encompasses nonlinear dimensionality reductions.  Feature selection is, by construction, linear, where the possible subspace is even further limited to linear spaces that are parallel to the feature space axes.  Lowering the feature dimensionality by feature selection can also aid in interpreting classification results.  At least it can shed some light on what features seem to matter mostly and possibly we can gain some insight into their interdependencies, for instance, by studying the coefficients of a trained linear classifier.  Aiming for a more interpretable classifier, we might even sacrifice some of the performance for the sake of a really limited feature set size.  Feature selection can also be used to select the right prototypes when employing the dissimilarity approach \cite{pekalska2006prototype}, as in this case every feature basically corresponds to all distances to a particular prototype.

\section{Evaluation}\label{sect:eval}

So how good are all these classification approaches really?  How can we decide for one feature set over the other?  How to compare two classifiers?  What prototypes work and which do not?  In Section \ref{sect:pre}, we already stated that the (true) error rate as defined in Equation \eqref{eq:err} is, in our context, the ultimate measure that decides which procedure is best: the lower the error rate, the better.
A first problem we face, the same one we encountered in dealing with the Bayes error and classifier---$\varepsilon^*$ and $C^*$, is that $p_{XY}$ is not available to us and so we are unable to determine the exact classification error for any procedure.  Like in building a classifier, we have to rely on the $N$ samples from $p_{XY}$ that we have and we can merely come to an estimated error rate.  But the situation is worse even.  Not only do we have to give an estimate of our performance measure based on these $N$ samples---the simplest of ways would be to just count the number of misclassifications on that set and divide that number by $N$.  Typically, we also have to use these same $N$ samples to train our classifier.  In many real-world settings there simply is not more data available or there is no time or money left to gather more.

We consider some alternatives to estimate an error rate (see also \cite{schiavo2000ten}), introduce so-called learning curves that give some basic insight into classifier behavior, mention overtraining and in what sense there is no single best classifier, and offer some further considerations when it comes to developing a complete classifier system.

\subsection{Apparent Error and Holdout Set}

A major mistake, which is still being made among users and practitioners of pattern recognition and machine learning, is that one simply uses all available samples to build a classifier and then estimates the error on these same $N$ samples.  This estimate is called the resubstitution or apparent error, denoted $\varepsilon^\mathrm{A}$ . The problem with this approach is that one gets an overly optimistic estimate.  The classifier has been adapting to these specific points with these specific labels and therefore performs particularly well on this set.  To more faithfully estimate the actual generalization performance of a classifier one would need a training set to train the classifier and a completely independent so-called test set\footnote{In particular settings, when dealing with transductive learning for instance \cite{vapnik1998statistical}, the inputs may be available and exploited in the training phase.  In that case, the training is still performed independent of the labels in the test set, which is the primary requirement.  The related setting of semi-supervised learning is briefly covered in Subsection \ref{sect:ssl}.} to estimate its performance.  The latter is also referred to as the holdout set.

In reality, we often have only a single set at our disposal, in which case we can construct a training and a holdout set by splitting the initial set in two.  But how do we decide on the sizes of these two sets?  We are dealing with two conflicting goals here.   We would like to train on as much data as possible, as this would typically give us the best performing classifier\footnote{Typically, yes, but classifiers can act counterintuitively and perform structurally worse with increasing numbers of labeled data in certain settings \cite{loog2012dipping}.\label{xx}}.  So the final classifier we would deliver, say, to a client, would be trained on the full initial set available.  But to get an idea of the true performance of this classifier---a possible selling point if low, we at least need some independent samples.  The smaller we take this set, however, the less trustworthy this estimate will be.  In the extreme case of one test sample, for instance, the estimate for error rate will always be equal to $0$ or $1$.  But adding data to the test set will reduce the amount of data in the training set, which removes us further from the setting in which we train our final classifier on the full set.  The following approaches, relying on resampling the training data, provide a resolution.

\subsection{Resampling Techniques}

We consider some resampling techniques that provide us with some possibilities to evaluate classifiers that, in real-world settings, are often more practical then using a holdout set.

\subsubsection[Leave-one-out and $k$-Fold Cross Validation]{Leave-one-out and $\bm{k}$-Fold Cross Validation}

Cross validation is an evaluation technique that offers the best of both worlds and allows us to both train and test on large data sets.  Moreover, when it comes to estimation accuracy, so-called leave-one-out cross validation is probably one of the best options we have.

The latter approach loops through the whole data set for all $i \in \{1,\ldots,N\}$.  At step $i$ the pair $(x_i,y_i)$ is used to evaluate the classifier that has been trained on all examples from the full set, except for that single sample $(x_i,y_i)$.  So we have a training set of size $N-1$ and a test set size of $1$.  Given that we want at least some data to test on, this is the best training set size we can have.  The test set size is almost the worst we can have, but this is just for this single step in our loop.  Going through all data available, every single sample will at some point act as a test set, giving us an estimated error rates $\varepsilon_i$ (all of value 0 or 1), which we can subsequently average to get to a better overall estimate
\begin{equation}\label{eq:looest}
  \varepsilon^\mathrm{LOO} = \frac{1}{N} \sum_{i=1}^N \varepsilon_i \, .
\end{equation}
This procedure is called leave-one-out cross validation and its resulting estimate the leave-one-out estimate \cite{efron1982jackknife, lachenbruch1968estimation, mclachlan2004discriminant}.

For computational reasons, e.g.\ when dealing with rather large data sets or classifiers that take long to train, one can consider to settle for so-called $k$-fold cross validation instead of its leave-one-out variant.  In that case, the original data set is split in $k$, preferably, equal sized sets or $k$ folds.  After this, the procedure is basically the same as with leave-one-out: we loop over the $k$ folds, which we consecutively leave out during training and which we then test on.  Leave-one-out is then the same as $N$-fold cross validation.

\subsubsection{Bootstrap Estimators}

Bootstrapping is a common resampling technique in statistics, the basic version of which samples from the observed empirical distribution with replacement.  Various bootstrap estimators of the error rate aim to correct the bias---that is, the overoptimism, in the apparent error.

One of the more simple approaches proceeds as follows \cite{efron1982jackknife, efron1983estimating}.  From our data set of $N$ training samples, we generate $M$ bootstrap samples of size $N$ and calculate the $M$ corresponding apparent error rates $\varepsilon_i^\mathrm{A}$ for our particular choice of classifier.  Using every time that same classifier, we also calculate the error $\varepsilon_i^\mathrm{T}$ rate on the total data set.  An estimate of the bias is now given by their averaged difference.
\begin{equation}\label{eq:bootstrapbias}
  \beta = \frac{1}{M} \sum_{i=1}^M \varepsilon_i^\mathrm{A} - \varepsilon_i^\mathrm{T} \, .
\end{equation}
The bias corrected version of the apparent error, and as such an improved estimate of the true error, is now given by $\varepsilon^\mathrm{A} - \beta$.

Various improvements upon and alternatives to this scheme have been suggested and studied \cite{efron1982jackknife, efron1983estimating, efron1997improvements}.  Possibly the best-known is the .632 estimator $\varepsilon^\mathrm{.632} = 0.368 \varepsilon^\mathrm{A} + 0.632 \varepsilon^\mathrm{O}$.  With the first term on the right-hand side the apparent error and the second term the out-of-bootstrap error.  The latter is determined by counting all the samples from the original data set that are misclassified and that are not part of the current bootstrap sample based on which the classifier is built.  Adding all these mistakes over all $M$ rounds and dividing this number by the total number of out-of-bootstrap samples, gives us $\varepsilon^\mathrm{O}$.

\subsubsection{Tests of Significance}\label{sect:teststats}

In all of the foregoing, it is of course important to remember that our estimates are based on random samples and so we will only observe an instantiation of a random variable.  As a result, to decide on anything like a significant difference in performances of two or more classifiers, we have to resort to statistical tests \cite{dietterich1998approximate}.   To get a rough idea of the standard deviation of any of our error estimates $\varepsilon^\mathrm{X}$ based on a test size of size $N_\mathrm{test}$, we can for instance use a simple estimator that can be derived as the result of an averaging of $N_\mathrm{test}$ Bernoulli trials:
\begin{equation}\label{eq:stdtest}
  \mathrm{std}[ \varepsilon^\mathrm{X}] = \sqrt{\frac{\varepsilon^\mathrm{X}(1-\varepsilon^\mathrm{X})}{N_\mathrm{test}}} \, .
\end{equation}

\subsection{Learning Curves and the Single Best Classifier}\label{sect:singlebest}

In Subsection \ref{sect:featcurv}, we briefly introduced feature curves, which give us an impression of how the error rate evolves with an increasing numbers of features.  We discussed the curse of dimensionality in this context.  It should be clear by now that also these feature curves can, like the error rate, only be estimated and to do so, one would typically apply the estimation techniques described in the foregoing.  Another, maybe more important curve that provides us with insight into the behavior of a classification method is the so-called learning curve \cite{cortes1993learning, duin2015, langley1988machine}.  The learning curve plots the (estimated) true error rate against the number of training samples.  To complete the picture, one typically also plots the apparent error in the same figure.

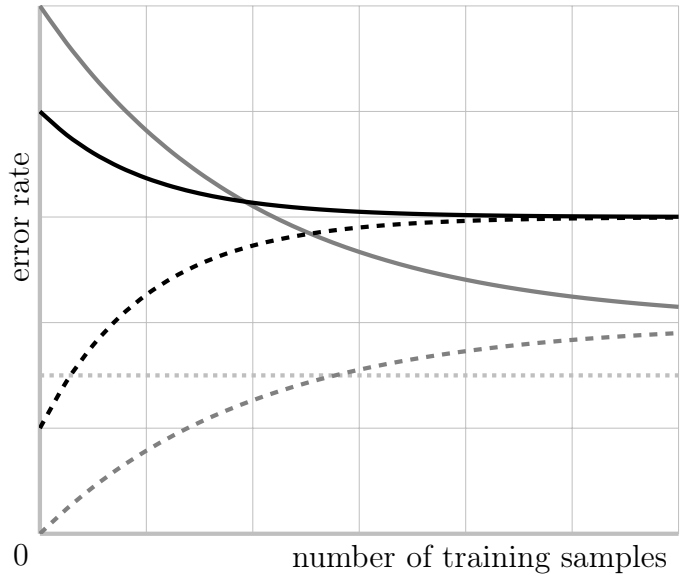
\begin{figure}
\center
\begin{tikzpicture}[scale=1.4]
\draw[step=1cm,lightgray,very thin] (0,0) grid (6,5);
\draw[ultra thick,lightgray,-] (0,0) -- (6,0);
\draw[black](6,0) node[anchor=north east] {number of training samples};
\draw[ultra thick,lightgray,-] (0,0) -- (0,5);
\draw (0,0) node[anchor=north east] {0};
\draw[black](0,3) node[anchor=south, rotate=90] {error rate};
\draw[ultra thick,lightgray,dotted] (0,1.5) -- (6,1.5);
\clip (0,0) rectangle (6,5);
\draw[ultra thick,gray,domain=0:6,smooth,variable=\x] plot ({\x},{(2+3*exp(-\x/2))});
\draw[ultra thick,gray,dashed,domain=0:6,smooth,variable=\x] plot ({\x},{(2-2*exp(-\x/2))});
\draw[ultra thick,black,domain=0:6,smooth,variable=\x] plot ({\x},{(3+exp(-\x))});
\draw[ultra thick,black,dashed,domain=0:6,smooth,variable=\x] plot ({\x},{(3-2*exp(-\x))});
\end{tikzpicture}
\caption{Stylized plots of the learning curves of two different classifiers.  Solid lines are the (estimated) true errors for different training set sizes, while the dashed lines sketch the apparent errors.  Black lines below to a classifier with relatively low complexity, while the gray lines illustrate the behavior of a more complex classifier.  The light gray, dotted, horizontal line is the Bayes error for the problem considered.}\label{fig:learn}
\end{figure}

Figure \ref{fig:learn} displays stylized learning curves for two classifiers of different complexity.  There are various characteristics of interest that we can observe in these plots and that reflect the typical behavior for many a classifier on many classification problems.  To start with, with growing sample size, the classifier is expected to perform better in terms of the error rate\footnote{See, however, footnote \ref{xx}.}.  In addition, for the apparent error we would typically observe the opposite behavior: the curve increases as it becomes more and more difficult to solve the classification problem for the growing training set\footnote{Clearly, there are exceptions.  For instance, the nearest neighbor classifier generally gives a zero error rate on the training set.}.  In the limit of an infinite amount of data points, both curves come together\footnote{Again exceptions apply.  Again the nearest neighbor is an example.  Statistical learning theory formally studies settings for which such consistency does apply.}: the more training data one has, the better it describes the general data that we may encounter at test time and the closer to each other true error and apparent error get. In fact, the gap that we see is an indication that the trained classifier focuses too much on specifics that are in the training but not in the test set.  This is called overtraining or overfitting. The larger the gap between true and apparent error is, the more overtraining has occurred.

From the way that both learning curves for one classifier come together, one can also glean some insight.  Classifiers that are less complex typically drop off more quickly, but also level out earlier than more complex ones.  In addition, the former converges to an error rate that is most often above the limiting error rate of the latter: given enough data, one can get closer to the Bayes error when employing a more complex classifier\footnote{This also shows that complex classifiers typically have a smaller bias than simple classifiers.}.  As a result, it often is the case that one classifier is not uniformly better than another, even if we consider the same classification problem.  It really matters what training set size we are dealing with and, when benchmarking the one classification method against the other, this should really be taken into account.  Generally, the smaller the training data set is, the better it is to stick with simple classifiers, e.g.\ using a linear hypothesis class and few features.

The fact that the best choice of classifier may depend not only on the type of classification problem we need to tackle, but also on the number of training samples that we have at our disposal, may lead one to wonder what generally can be said about the superiority of one classifier over the other. Wolpert \cite{wolpert1992connection, wolpert1996lack} (see also \cite{duda2001pattern}) made this question mathematically precise and demonstrated that for this and several variations of this question the answer is that, maybe surprisingly, there does not exist such distinctions between learning algorithms.  This so-called no free lunch theorem states, very roughly, that averaged over all classification problems possible, there is no one classification method that outperforms any other.  Though the result is certainly gripping, one should interpret it with some care.  It should be realized, for instance, that among all possible classification problems that one can construct, there probably are many that do not reflect any kind of realistic setting.  What we can say, nevertheless, is that generally there is no single best classifier.

Finally, a learning curve may give us an idea of whether gathering more training data may improve the performance.  In Figure \ref{fig:learn}, the classifier corresponding to the black curves can hardly be improved, even if we add enormous amounts of additional data.  The other classifier, the gray curves, can probably improve a bit, reaching a slightly lower error rate when enlarging the traning set.

\subsection{Some Words about More Realistic Scenarios}

In real-world applications, designing and building a full classifier system will often be a process in which one may consider many feature representations, in which one will try various feature reduction schemes, and in which one will compare many different types of classifiers.  On top of all that, there might be all kinds of preprocessing steps that are applied to the data (and that are not explicitly covered in this chapter). Working with images or signals for instance, one can perform various types of enhancement, smoothing, and normalization techniques that may have positive or negative effect on the performance of our final classifier.

A real problem in all this is that it is difficult to dispose of a truly independent test set.  Unless one has a massive amount of labeled training data, one easily gets into the situation that data that is also going to be used for evaluation leaks into the training phase.  The estimated test errors are therefore overly optimistic and more so for complex classifiers than for simple ones.  In the end, the result of this is that we may end up with a wrongly trained classifier, together with an overly optimistic estimate of its performance.

Let us consider some examples where things go wrong.
\begin{itemize}

\item

A very simple instance is where one has decided, at some point, to use the $k$ nearest neighbor classifier. The only thing that remains to be done is finding the best value for $k$ and one decides to determine it on the basis of the performance for every $k$ on the test set.  It may seem like a minor offense, but often there are many of such choices: the best number of features, the number of nodes in a layer of a neural network, the free parameters in some of the kernels, etc. (cf. \cite{levine2009submissions} and, in particular point 7 in the list).

\item

Here is an example where it is maybe more difficult to see that one may have gone wrong.  We decide to set up everything in a seemingly clean way. We prefix all classifiers that we want to study, all the feature selection schemes that we want to try, decide beforehand on all the kernels we want to consider, and all classifier combining schemes that we may want to employ.  This gives a finite number of different classification schemes that we then compare based on cross validation.  In the end, we then pick the scheme that provides the best performance.  Even though this approach is actually fairly standard, again something does go wrong here.  If the number of different classification schemes that we try out in this way gets out of hand, and it easily does, we still run the risk that we pick an overtrained solution with a badly biased estimate for its true error rate, especially when dealing with small training sets (cf.\ \cite{braga2004cross, isaksson2008cross, schaffer1993selecting}).

\item

Even more complicated issues arise when multiple groups work on a large and challenging classification task.  Nowadays, there are various research initiatives in which labeled data is provided publicly by a third party on which researchers can work simultaneously and collaboratively, but also in competition with each other.  The flow of information and, in particular, the possibly indirect leakage of test data becomes difficult to oversee, let alone that we can easily correct for it when providing error estimates and corresponding confidence intervals or the like.  How does one, for instance, correct for the fact that one's own method is inspired by some of the results by another group one has read about in the research literature?  Though some statistical approaches are available that can alleviate particular problems \cite{dwork2015reusable, dwork2015generalization}, it is safe to say that there currently is no generally applicable solution---if such at all exists.

\end{itemize}

Now the above primarily pertains to evaluation.  In real scenarios, we of course also have to worry about the reproducibility and replicability of our findings.  Otherwise, what kind of science would this be?  Clearly, these are all issues that in one way or the other also play a significant role in other areas of research.  In general, it turns out, however, that it is difficult to control all of these aspects and that mistakes are made, mostly unwittingly but in some case possibly even knowingly.  For some potential, more or less dramatic  consequences, we refer to the following good reads: \cite{duin1994superlearning, fanelli2011negative, ioannidis2005most, leek2015statistics, moonesinghe2007most, nissen2016publication}.

\section{Regularization}\label{sect:reg}

Regularization is actually a rather important yet relatively advanced topic in supervised learning \cite{christianini2000support, poggio2003mathematics, scholkopf2002learning, vapnik1998statistical, wahba1981constrained} and unfortunately we are going to be fairly brief about it here.

The main idea of regularization is to have a means of performing complexity control.  As we have seen already, classifier complexity can be controlled by the number of features that are used or through the complexity of the hypothesis class and, in a way, regularization is related to both of these.  One of the well-known ways of regularizing a linear classifier is by constraining the already limited hypothesis space further.  This is typically done by restricting the admissible weights $w$ of the linear classifier to a sphere with radius $t > 0$ around the origin of the hypothesis space, which means we solve the constraint optimization problem
\begin{equation}\label{eq:optreg}
\begin{split}
h^* & = \operatorname*{argmin}_{(w,w_\circ) \in \mathbb{R}^{d+1}} \sum_{i=1}^N   \ell(w^\top x_i + w_\circ,y_i) \\
& \text{subject to} \mbox{ } || w ||^2 \le t \, .
\end{split}
\end{equation}
A formulation that is essentially equivalent is constructed by including the constraint directly into the objective function:
\begin{equation}\label{eq:optreg2}
h^* = \operatorname*{argmin}_{(w,w_\circ) \in \mathbb{R}^{d+1}} \sum_{i=1}^N   \ell(w^\top x_i + w_\circ,y_i) + \lambda || w ||^2 \, ,
\end{equation}
where $\lambda > 0$ is known as the regularization parameter.  The regularization is stronger with larger $\lambda$.

This procedure is the same as the one used in classical ridge regression \cite{hoerl1970ridge} and effectively stabilizes the solution that is obtained. The effect of regularization is that the bias of our classification method increases, as we cannot reach certain linear classifiers anymore due to the added constraint.  At the same time, the variance in our classifier estimates decreases due to the constraint (which is another way of saying that the classifier becomes more stable).  In the average, with a small to moderate parameter $\lambda$, the worsening in performance we may get because of the increased bias is amply compensated with an improvement in performance due to the reduced variance, in which case regularization will lead to an improved classifier.  If, however, we regularize too strongly, the bias will start to dominate and pull our model too far away from any reasonable solution at which point the true error rate will start to increase again.

A basic explanation of the effects of this so-called bias-variance tradeoff can already be found in the earlier mentioned work of Hoerl and Kennard \cite{hoerl1970ridge}.  The phenomenon can be seen in various guises and its importance has been acknowledged early on in statistics and data analysis \cite{wahba1979smoothing, wahba1981constrained}.  A more explicit dissection of the bias-variance tradeoff, in the context of learning methods, was published in \cite{geman1992neural}.   The more complex a classifier is, the higher the variance we are faced with when training such model, and the more important some form of regularization becomes.

Equations \eqref{eq:optreg} and \eqref{eq:optreg2} only consider the most basic form of regularization.  There are many more variations on this theme. Among others, there are regularizers with built-in feature selectors \cite{tibshirani1996regression} and regularizers that have deep connections to our earlier discussed kernels \cite{girosi1995regularization, smola1998kernel} .

\section{Variations on Standard Classification}\label{sect:disc}

We have introduced and discussed the main aspects of supervised classification.  In this last section, we like to review some slight variations to this basic learning problem. Though basic, there are very many decision problems that can actually be cast into a classification problem.  Nevertheless, in reality, one may often be confronted with problems that still do not completely fit this restricted setting.  Some of these we cover here.

\subsection{Multiple Instance Learning}

In particular settings, it is more appropriate or it simply is easier to describe every object $o_i$, not with a single feature vector $x_i$, but with a set of such feature vectors.  This approach is, for example, common in various image analysis tasks, in which a set of so-called descriptors, i.e., feature vectors that capture the local image content at various locations in the image, act as the representation of that image.  Every image, in both the training and the test set, is represented by such a set of descriptors and the goal is to construct a classifier for such sets.  The research area that studies approaches applicable to this setting, in which every object can be described with sets of feature vectors having different sizes, but where the feature vectors are from the same measurement space, is called multiple instance learning.  A large number of classification routines have been developed for this specific problem, which range from basic extensions of classifiers from the supervised classification domain by means of combining techniques, via dissimilarity-based approaches, to approaches specifically designed for the purpose of set classification \cite{carbonneau2016multiple, cheplygina2015multiple, li2013multiple, maron1998framework, zhou2004multi}.  The classical reference, in which the initial problem has been formalized, is \cite{dietterich1997solving}.

\subsection{One-class Classification, Outliers, and Reject Options}

There are various problems where it is difficult to find sufficient examples of one of the classes, because they are very difficult to find or simply occur very seldom.  In that case, one-class classification might be of use.  Instead of trying to solve the two-class problem straightaway, it aims to model the distribution or support of the oft-occurring class accurately and based on that decides which points really do not belong to that class and, therefore, will be assigned to the class of which little is known \cite{scholkopf2001estimating, tax1999support}.  Such techniques have direct relations to approaches that perform outlier or novelty detection in data and data streams \cite{chandola2009anomaly, markou2003novelty} in which one aims to identify objects that are, in some sense, far away from the bulk of the data.

The more a test data point is an outlier, the less training data will be present in its vicinity and, therefore, the less certain a classifier will be in assigning the corresponding object to one or the other class.  Consequently, outlier detection and related techniques are also used to implement so-called reject options \cite{chow1970optimum}. These aim to identify points for which, say, $p_X$ is small and any automated decision by the classifier at hand is probably unreliable. In such case, the ultimate decision may be better left to a human expert. We might, for instance, be dealing with a sample from a third class; something that our classifier never saw examples of.  This kind of rejection is also referred to as the distance reject option \cite{dubuisson1993statistical}.  A second option is ambiguity rejection, in which case the classifier rather looks at $p_{Y|X}$ and leaves the final decision to a human expert if (in the two-class case) the two posteriors are very close to each other, i.e., approximately $\tfrac{1}{2}$ \cite{dubuisson1993statistical}.  For ambiguity and distance rejection, one should realize that both an erroneous decision by the classifier and deploying a human expert come with their own costs. One of the main challenges in the use of a reject option is then to trade these two costs off in an optimal way.

\subsection{Contextual Classification}

Contextual classification has already been mentioned in Subsection \ref{sect:mcs} on multiple classifier systems.  In these contextual approaches, samples are not classified in isolation, but they may have various types of neighborhood relations that can be exploited to improve the overall performance.  The classical approach to this employs Markov random fields \cite{besag1986statistical, li2009markov} and specific variations to those techniques like conditional random fields \cite{lafferty2001conditional}.  The earlier mentioned methods using classifier combining techniques \cite{cohen2005stacked, loog2002supervised, loog2004supervised} are often more easily applicable and can leverage the full potential of more general classification methodologies.  As already indicated, in Subsection \ref{sect:mcs} as well, the latter class of techniques seems to become relevant again in the context of nowadays deep learning approaches.

\subsection{Missing Data and Semi-supervised Learning}\label{sect:ssl}

In many real-world setting, missing data is a considerable and reoccurring problem.  In the classification setting this means that particular features and/or class labels have not been observed.  Missing features can occur because of the failure of a measurement apparatus, because of human non-response, or because the data was not recorded or got accidentally erased.  There are various ways to deal with such deletions, which is a topic thoroughly studied in statistics \cite{little2014statistical}.

The case of missing labels can have additional causes. It may simply have been too expensive to label more data or additional input data has been collected afterwards to extend the already available data, but the collector is not a specialist that can provide the necessary annotation.  The case of missing label data is known within pattern recognition and machine learning as semi-supervised learning \cite{chapelle06b, zhu08a}.  Also for this problem, which has been studied for over 50 years already, many different techniques have been developed.  Though maybe more in a theoretical sense, there is still no completely satisfactory and practicable solution to the problem\footnote{In a way, this can probably also be said about many of the other problems discussed.}. One of the major issues is the question to what extent one can guarantee that a supervised classifier can indeed be improved by taking all unlabeled data into account as well \cite{Krijthe2017, li2011towards, loog2010constrained, loog2016contrastive}.

\subsection{Transfer Learning and Domain Adaptation}\label{sect:tlda}

For various kinds of reasons, the distribution of the data at training time can be rather different from that at test time.  Examples are medical devices that are trained on samples (subjects) from one country, while the machine is also deployed in another country.  More generally, machines and sensors suffer from wear and tear and, as a result, measurement statistics at a later point in time may not really match their distribution at time of training.  Depending on what can be assumed about the difference of the two domains or, as they are often referred to more specifically, the source and target, particular approaches can be employed that can alleviate the discrepancy between them \cite{pan2010survey, quinonero09}.  The areas of domain adaptation and transfer learning study techniques for these challenging settings.  Depending on the actual transfer that has to be learned, more or less successful approaches can be identified to tackle these problems.

\subsection{Active Learning}

The final variation on supervised classification is actually concerned with regular supervised classification.  The difference, however, with the main setting discussed throughout this chapter is that active learning sets out to improve the data collection process.  It tries to answer various related questions, one of which is as follows. Given that we have a large number of unlabeled samples and a budget to label $N$ of these samples, what instances should we consider for labeling to enable us to train a better classifier than we would be able to in case we would rely on random sampling\footnote{The latter of which is the de facto standard.}?  So can we in a more systematic way collect data to be labeled, such that we quicker come to a well-trained classifier?

The problem formulation has direct relations to sequential analysis \cite{wald1973sequential} and optimal experimental design \cite{fedorov1972theory}.  Overviews of current techniques can be found in \cite{cohn1996active}, \cite{settles2010active}, and \cite{yang2016benchmark}.  One of the major issues in active learning is that the systematic collection of labeled training data typically leads to a systematic bias as well. Correcting for this seems essential \cite{beygelzimer2009importance} (see also \cite{loog2016empirical}).  In a way, it points to a problem one will more generally encounter in practical settings and which directly relates to some of the issues indicated in Subsection \ref{sect:tlda}: one of the key assumptions in supervised classification is that the training and test set consist of i.i.d.\ samples from the same underlying problem defined by the density $p_{XY}$.  In reality, this assumption is most probably violated and care should be taken.

\section*{Acknowledgements}

Many thanks go to Mariusz Flasi\'{n}ski (Jagiellonian University, Poland) and a second, anonymous reviewer for their critical, yet encouraging appraisal of this overview.  Their comments helped to improve the exposition of the chapter and to drastically reduce the number of spelling and grammar glitches.

\bibliographystyle{plain}
\bibliography{PRchap}

\end{document}